\documentclass[letterpaper, 10 pt, conference]{ieeeconf} 

\IEEEoverridecommandlockouts               

\usepackage{cite}

\usepackage{amsmath,amssymb,amsfonts}
\usepackage{algorithmic}
\usepackage{graphicx}
\usepackage[caption=false,font=footnotesize]{subfig}
\usepackage{textcomp}
\usepackage{balance}
\usepackage{lipsum}
\usepackage{array}
\usepackage{arydshln}
\usepackage{todonotes}
\usepackage{threeparttable}
\usepackage{siunitx}
\newcolumntype{C}[1]{>{\centering\let\newline\\\arraybackslash\hspace{0pt}}m{#1}}

\usepackage{xcolor}

\title{\LARGE \bf
Force Sensing for Wearable Human-Robot Interfaces via Fluidic Innervation
}

\author{Noah Rubin$^1$, Ava Schraeder$^{2,*}$, Hrishikesh Sahu$^{2,*}$, Thomas C. Bulea$^1$, and Lillian Chin$^2$
\thanks{*These authors contibuted equally to this work.}
\thanks{$^{1}$Rehabilitation Medicine Department, National Institutes of Health (NIH) Clinical Center, Bethesda, Maryland, USA.}
\thanks{$^{2}$Department of Electrical and Computer Engineering, University of Texas at Austin, Austin, Texas, USA.}
}
\begin{document}
\maketitle
\thispagestyle{empty}
\pagestyle{empty}


\begin{abstract}
Mechanically characterizing the human-machine interface is essential to understanding user behavior and optimizing wearable robot performance. This interface has been challenging to sensorize due to manufacturing complexity and non-linear sensor responses. Here, we measure human limb-device interaction via fluidic innervation, creating a 3D-printed silicone pad with embedded air channels to measure forces. As forces are applied to the pad, the air channels compress, resulting in a pressure change measurable by off-the-shelf pressure transducers. We demonstrate in benchtop testing that pad pressure is highly linearly related to applied force ($R^2 = 0.998$) and confirmed strong linear relationships to isometric knee torque in a clinical dynamometer with strategic pad placement. We built on these idealized settings to test pad performance in more unconstrained settings, including during cyclic dynamic and stepwise isometric bicep curls. Finally, we integrated the sensor into a lower-extremity robotic exoskeleton and recorded pad pressure during repeated squats with the device unpowered. Pad pressure tracked squat phase and overall task dynamics consistently. Collectively, our preliminary results suggest fluidic innervation is a readily customizable sensing modality with high signal-to-noise ratio and temporal resolution for capturing human-machine interaction. In the long-term, this modality may provide an alternative real-time sensing input to control / optimize wearable robotic systems and to capture user function during device use. 
\end{abstract}

\section{Introduction}

\begin{figure}[t]
  \centering
  \includegraphics[width=0.99\linewidth]{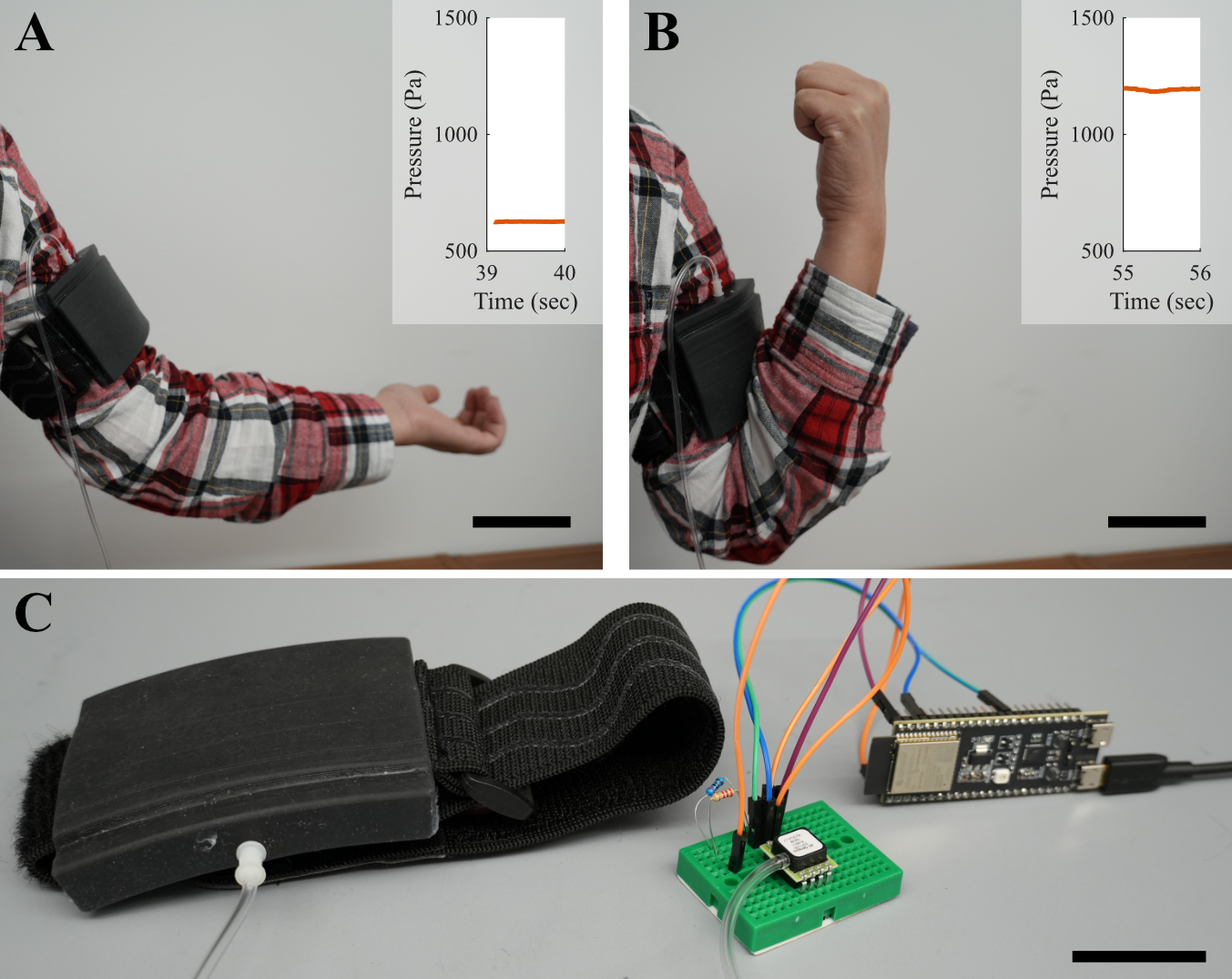}
  \caption{Demonstration of a person wearing the sensorized pad at (A) rest and (B) full elbow flexion. The insets show representative data of the reported pad pressure. (C) Overview of the sensorized pad system. From left to right, the black fluidically innervated pad on a strap. The pad is connected by a white barb and silicone tubing to an off-the-shelf pressure transducer on a breadboard. The transducer connects to an ESP-32 microcontroller which streams the data to the main computer. All scale bars are 5 \unit{\centi\metre}} 
  \label{fig:overview}
  \vspace{-5mm}
\end{figure}

Wearable robotic devices are becoming increasingly popular as a method to enhance and measure human movement. This includes both rigid exoskeletons that exert a force on the human body to assist and/or rehabilitate movement \cite{rosenWearable2019}, as well as soft robotic haptic interfaces, compression therapies, and dynamically-changing fashion \cite{zhuSoft2022}. A fundamental design challenge persists across these applications: what should the mechanical interface between human and machine look like? This interface is defined by the forces and torques applied by the human against their device (and vice versa). Thus, mechanically characterizing the human-device interaction can provide useful information, such as movement intent and force output, as well as providing insight into fundamental aspects of user motor control and even pathophysiology \cite{adamczyk2023wearable}.

For wearable robots to  dynamically respond to changes in both the external environment and human output, they must have real-time feedback of the mechanical interactions between the user and their machine. However, current approaches tend to instrument solely the human \textit{or} the robot, limiting our understanding of the state of the complete system \cite{babicChallenges2021}. Current human-centered signals like electromyography (EMG) often have low signal to noise ratios (SNRs) or slowly changing dynamics \cite{zhang2017human}, making it challenging for robot controllers to adapt to user behavior in real time. Robot-centered sensors like inertial measurement units (IMUs), reaction torque sensors, motor-generated voltages, and Bowden cable lengths measure the human by proxy \cite{leeControlling2016,lunenburgerBiofeedback2004}, which can introduce delays and uncertainty between the real and measured signals. This complicates real-time control, as evidenced by recent efforts that use machine learning to combine multiple proxy measurements to achieve gait assistance \cite{molinaroTaskagnostic2024,luoExperimentfree2024}.

\begin{figure}[t]
  \centering
  \includegraphics[width=0.99\linewidth]{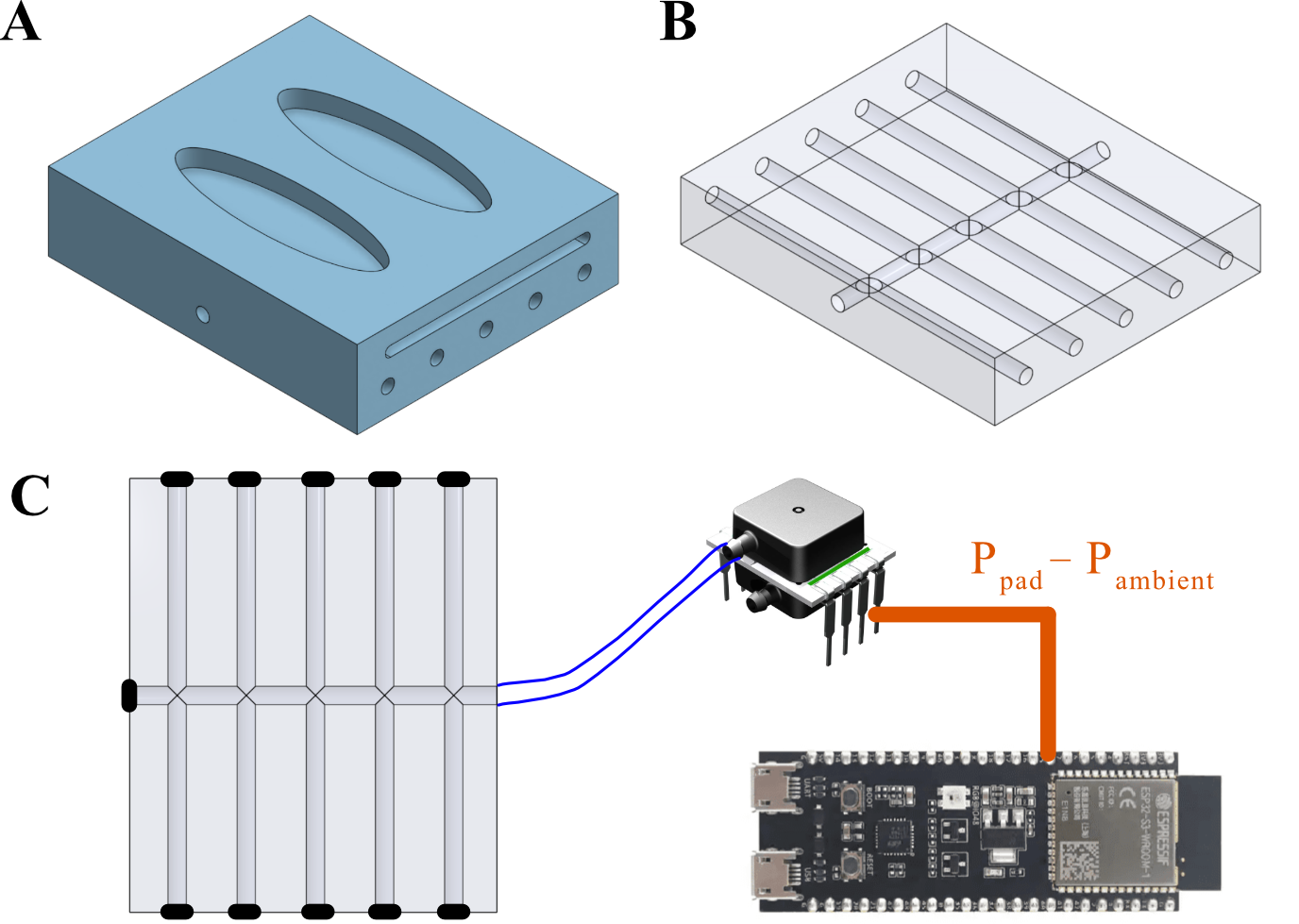}
  \caption{CAD render and system overview of the fluidically innervated pad. (A) Render of the entire pad with upper slot for strap attachment. (B) Transparent view without the strap slot showing  interconnected channel structure yielding one bulk pressure reading. (C) Schematic of connection to readout electronics. All openings are sealed (black) except one connected via tubing (blue) to a differential pressure transducer measuring pad pressure relative to ambient air, with data recorded by a microcontroller.}
  \label{fig:padDesign}
  \vspace{-5mm}
\end{figure}

Previous systems have measured human-robot interaction forces by adding force sensors into pads, straps and other mechanical attachments within the robot body. Force-sensitive resistors (FSRs) are the most popular sensors due to their low cost and straightforward electronics \cite{tamez2015real,bessler2019prototype}. However, their non-linear response curve and significant imprecision limit their utility in applications requiring fast and accurate readings \cite{velasquez2019error, choi2018compact, rathore2016quantifying}. Optoelectronic and pneumatic solutions offer higher accuracy, but at the cost of multi-step casting for fabrication \cite{lenzi2011measuring, langlois2020investigating,wang2021differential}. 3D printing offers a potential way to ease fabrication difficulties, but current capacitive-based solutions require fine measurements of small capacitance changes \cite{langlois2021integration}. A sensing solution that is simple to manufacture, has high time resolution, and is adaptable across human interfaces has the potential to facilitate more robust real-time and dynamic control of wearable robotics.

In this work, we introduce fluidic innervation (FI) as a promising method for mechanical sensing of human-machine interaction in wearable devices (Fig.~\ref{fig:overview}). FI is a sensorization technique that adds force sensing to a 3D printed structure by embedding internal air channels at ambient air pressure \cite{truby2022fluidic}. Compression of the structure therefore compresses the internal air channels, producing measurable changes in pressure. Because FI leverages 3D printed soft materials with off-the-shelf components, application-specific designs can be rapidly fabricated. Although FI has been demonstrated to have a high SNR and time resolution in soft robotic grippers \cite{zhang2024embedded,shang2025forte}, it has not been tested in wearable applications. This is a significant technical gap, as human-centered applications innately have more variability than pure robotic applications. The rigid mounting used in prior work is inappropriate for wearability, making it unclear if the sensor performance will generalize to human-robot sensing.

First, we introduce the design of FI pads that can be worn on a human limb. Next, we characterize performance through off-human mechanical compression testing and on-human testing during isometric torque generation with a clinical dynamometer. Finally, we demonstrate the pads in less constrained tasks, including isometric and dynamic bicep curls as well as squats. Our results validate FI–based sensing of human–machine interaction across multiple activities, motivating its potential use as a measurement of human-device behavior and a control input in actuated robotic devices.

\section{Methods}

\subsection{Fabrication}
We adapted FI to the human context by designing a soft pad ($75 \times 64 \times 14$ \unit{mm}) that can be comfortably placed on the human body while minimally interfering with movement. There are 3 major components to the wearable system: the FI pad, readout electronics, and attachment to the human body (Fig.~\ref{fig:overview}C). 
The pad design and readout electronics were adapted from our prior work developing sensorized robots \cite{truby2022fluidic,shang2025forte}. Unlike prior work, where a structure contained multiple internal channels to capture localized surface information \cite{zhang2024embedded}, the pad in our work has only one internal channel. This single channel design allows us to get a single pressure measurement that represents an overall bulk interaction force between the human and the pad. The channel is designed as a series of interconnected tubes, forming a single cavity that covers most of the pad area (Fig.~\ref{fig:padDesign}A-B). We chose a channel diameter of 3.2 \unit{\milli\metre}) for ease of printability (Fig. S1 of \cite{truby2022fluidic}). 

Each pad was fabricated by a resin 3D printer using biocompatible Silicone 40 (Formlabs). This material was chosen because it is one of the softest commercial resins available, promoting maximum comfort. After printing, each channel was flushed with isopropyl alcohol and compressed air to ensure it was empty before sealing all but one end. This open end was connected to an off-the-shelf pressure transducer (AllSensors 10 in. H$_2$O) to measure the pressure within the channels of the pad (Fig.~\ref{fig:padDesign}C). Channel ends were sealed with Sil-poxy and cyanoacrylate glue, while the open channel end was connected to the pressure transducer via silicone tubing with plastic barbs. We used differential pressure transducers, but compare the measured pressure to ambient air (i.e., one port disconnected). This setup means that if the channels were bent but not compressed, the channel's volume of air would increase and a "negative" pressure would be reported \cite{truby2022fluidic}. The pressure transducer voltage was read via I2C by an ESP32, which then communicated via Bluetooth to a computer for data recording (sampled at 50 \unit{\hertz}, well above typical task dynamics \cite{winter2009biomechanics}).

For consistency, we used the same 77.8 \unit{\gram} pad for all experiments. However, we note that 3D printing makes it easy to tailor pads to fit a specific person.  We showcase this customization by printing the pads with various ``slot'' geometries to accommodate different straps for each application while the channel layout within each pad remained the same. Three different strap attachments were used for our in-human applications: clinical dynamometer (Sec.~\ref{sec:dynamometer}), bicep curls (Sec.~\ref{sec:arms}), and squats (Sec.~\ref{sec:squats}). 

\subsection{Characterization}

\subsubsection{Off-board Mechanical Testing}
\begin{figure}[!t]
  \centering
  \includegraphics[width=0.99\linewidth]{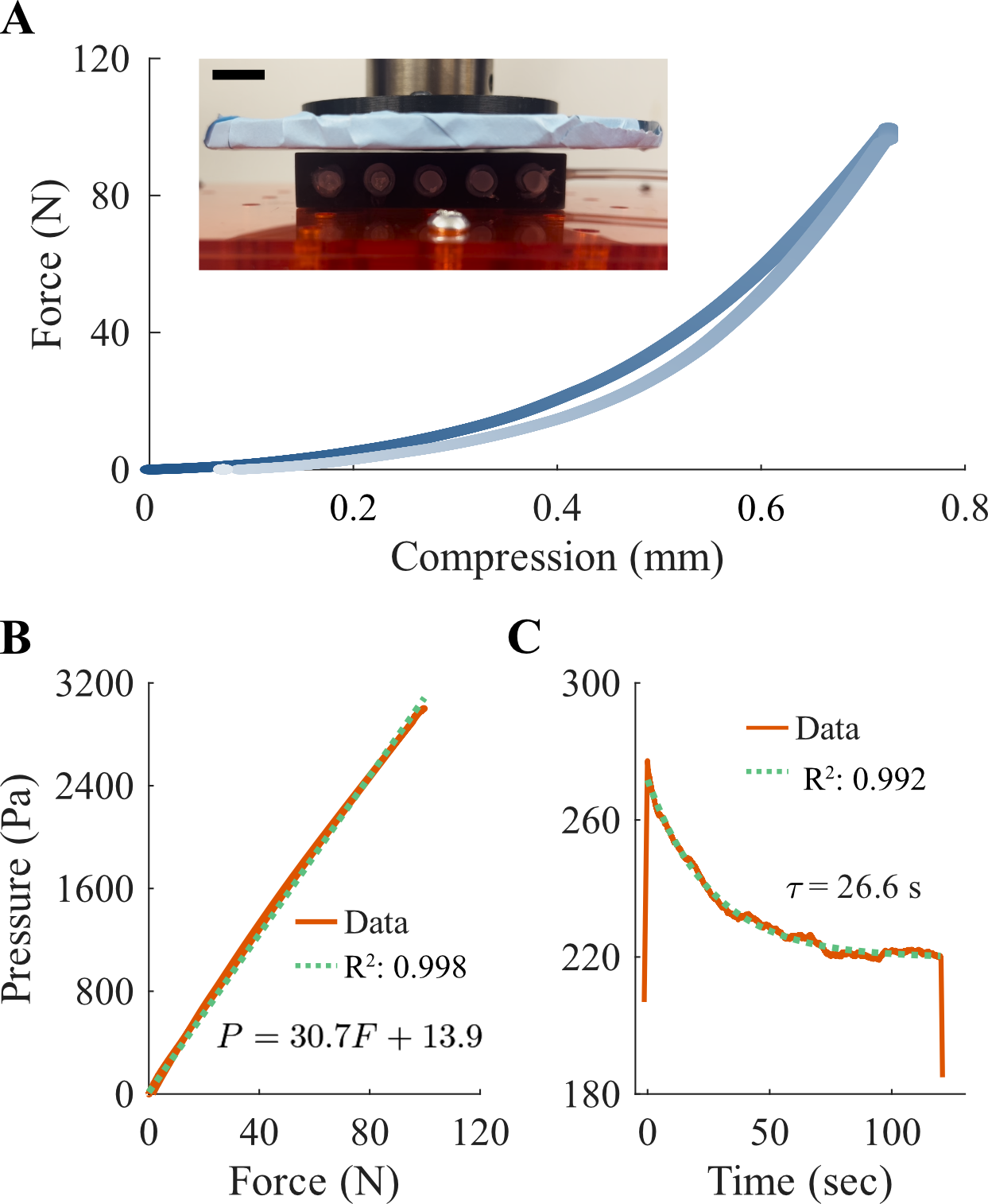}
  \caption{Characterization of sensorized pad using mechanical testing machine. (A) Mean compression distance vs. force across three trials (darker loading and lighter unloading). The max  deviation was 0.15 \unit{\newton} and $<10^{-2}$ \unit{\milli\metre} (not visible). The inset shows the compression test setup (scale bar 1 \unit{\centi\metre}). (B) Mean force vs. sensorized pad pressure was highly linear ($R^2$ 0.998, max SD 32 \unit{\pascal}, not visible). (C) Exemplar trial in which the pad was compressed until 20 \unit{\newton}, after which the position was held for 2 minutes. Stress relaxation in pressure readings were observed due to viscoelastic material properties, with an exponential decay time constant $\tau$ of 26.6 \unit{\second}.}
  \label{fig:instron}
  \vspace{-5mm}
\end{figure}

To validate the pads' ability to accurately measure applied force, we conducted compression tests on the pads using a mechanical testing machine (Shimadzu AGS-X). Each pad was placed under a compression plate to measure performance under bulk deformation, contrasting against the mechanical testing machine's ground truth of applied force vs. compression distance (Fig.~\ref{fig:instron}A). We downsampled and interpolated the 100 \unit{\hertz} data to match the sensorized pad's 50 \unit{\hertz} rate, and correlated force to pad pressure (Fig.~\ref{fig:instron}B). A preliminary test determined that the pressure transducer saturated at 113 \unit{\newton} of compression, hitting a measurement limit of 3114 \unit{\pascal}. To remain within sensing range, we compressed the pad at 1 \unit{\milli\metre/\second} up to 100 \unit{\newton}, held position for 10 \unit{\second} and unloaded at the same rate. We performed three trials and reported mean and maximum standard deviation (SD).

Sensorized pad pressure was highly linearly correlated with ground truth force (Fig.~\ref{fig:instron}B, $P=30.7F+13.9$,  $R^2=0.998$), demonstrating that measuring the embedded air channels' pressure accurately match the deformation of the overall pad. Silicone is a viscoelastic material, which means that it has a non-linear relationship between stress and strain. This is observed by the curved force vs. compression graph in Fig.~\ref{fig:instron}A, showing the non-linear response of the silicone material. Since Fig.~\ref{fig:instron}B shows a linear relationship, we see that our measurement technique does not add \textit{additional} non-linearity to the already non-linear silicone material. We can thus trust that our pressure measurements accurately follow the overall deformation of the pad structure. 

\begin{figure*}[t]
  \centering
  \includegraphics[width=\textwidth]{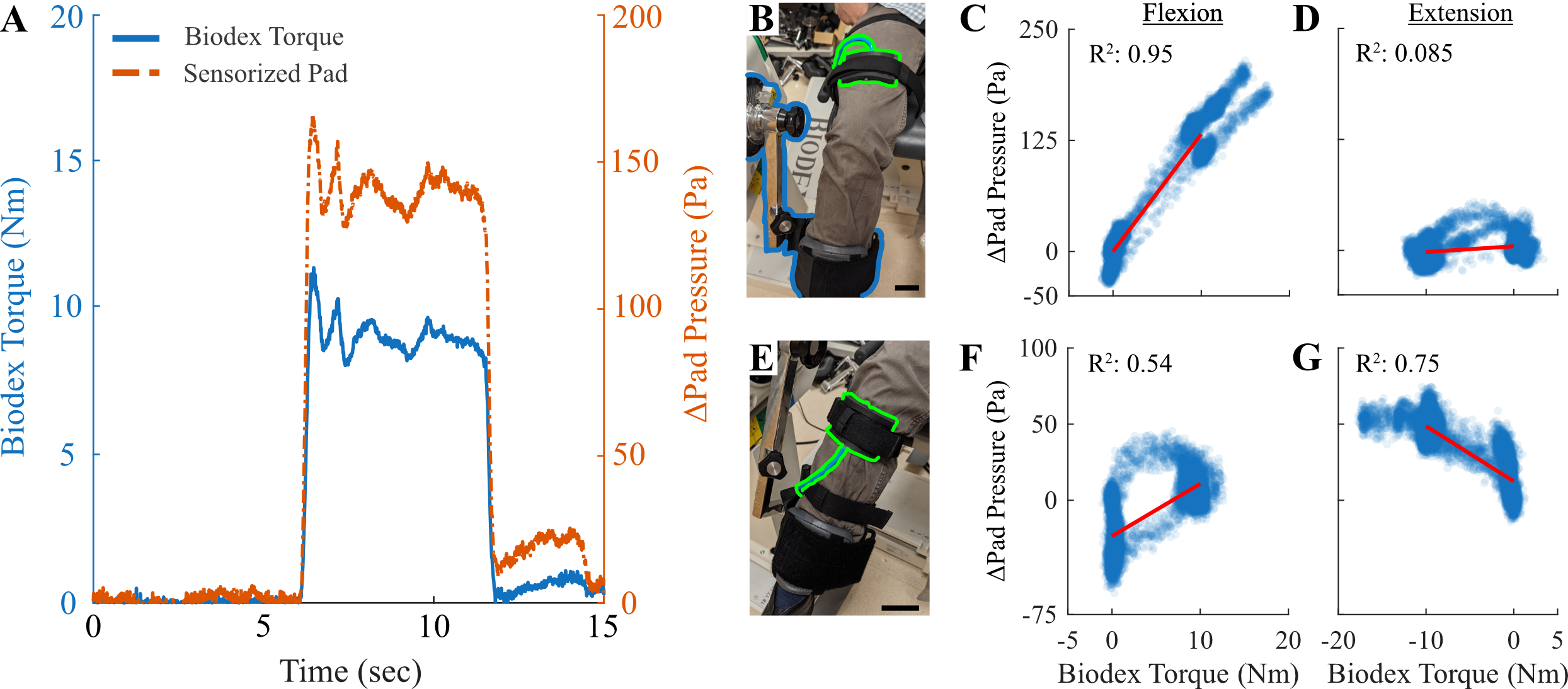}
  \caption{Characterization of sensorized pads through on-human dynamometer testing. Participants repeated isometric volitional knee torque output to 10 Nm while dynamometer torque and pad pressure were recorded simultaneously. Dynamometer is highlighted in blue while the fluidically innervated pad is highlighted in green (A) Exemplary time series data from knee flexion with (B) the fluidically innervated pad above the knee. (C-D) Correlation results between the measured pressure of the above-knee pad and the measured dynamometer torque during flexion and extension. (E) Tests were also performed with the fluidically innervated pad below the knee. (F-G) Correlation results between the measured pressure of the below-knee pad and the measured dynamometer torque during flexion and extension. Scale bars represent 5 \unit{\centi\metre}.
}
  \label{fig:biodex}
  \vspace{-3mm}
\end{figure*}

Another consequence of the pads being made from a viscoelastic material is \textit{stress relaxation}, an observed decrease in stress when a material is compressed to a consistent strain. In our sensors, this shows as a decrease in measured pressure when the pad is held to a consistent level. To characterize the decay response for our pads, we conducted a step compression test, similar to our prior FI work (Fig. 3 of \cite{truby2022fluidic}). The pad was compressed until 20 \unit{\newton} of force was measured. The position of the compression plate was then held for two minutes before unloading the pad (Fig.~\ref{fig:instron}C). We fit an exponential decay curve to this response, yielding a time constant of $\tau = 26.6$ \unit{\second}. The wearable pads are intended to capture forces during human movements lasting 2-3 seconds at most, indicating this decay will have minimal effect over the course of the movement ($< 10\%$ change in reading). Overall, these results demonstrate FI pads accurately measure applied forces. 

\subsubsection{On-Human Testing with Clinical Dynamometer}
\label{sec:dynamometer}

To demonstrate the pads' feasibility for capturing isolated user-device interaction, we contrasted FI measurements against those of a clinical dynamometer (Biodex System 4 Pro) in a controlled seated task. A single healthy individual (male, age 42 years) produced volitional isometric knee torque with their knee fixed at 45\unit{\degree} (Fig.~\ref{fig:biodex}). Four conditions were tested involving two pad locations (above and below knee (Fig.~\ref{fig:biodex}B,E)) and two torque directions (knee extension and flexion). The experiment order was conducted with the pad above the knee, followed by below the knee, and within each location, extension followed by flexion. Within each condition, four trials were completed, with 5 \unit{\second} of rest, 5 \unit{\second} holding torque at 10 \unit{\newton\metre}, and another 5 \unit{\second} of rest (Fig.~\ref{fig:biodex}A). The user was provided real time feedback of their torque output. Dynamometer torque and pad pressure were acquired as synchronized analog inputs (National Instruments USB-6211, sample rate 1000 Hz) into MATLAB (Mathworks, Natick, MA). After the experiment, torque and pressure data were converted from \unit{\volt} to \unit{\newton\metre} and \unit{\pascal}, respectively. Data were offset by the mean during the initial rest from 0.5 to 2 \unit{\second} to compute relative changes during the activity, and lowpass filtered to isolate task dynamics (6 \unit{\hertz} cutoff \cite{winter2009biomechanics}). Torque and pad data across all trials for each condition were linearly correlated, and $R^2$ values for the model were extracted. 

The user successfully modulated torque and held a steady-state value of approximately 10 \unit{\newton\metre}. Strikingly, pad pressure yielded both a very high signal to noise ratio from rest to 10 \unit{\newton\metre} and a very fast signal response to changes in volitional torque (Fig.~\ref{fig:biodex}A) in the scenario with the clearest interaction between pressure and torque output (pad above the knee during flexion as in Fig.~\ref{fig:biodex}B). Linear fits between torque and pad pressure varied by location and torque direction. Notably, the pad above the knee (Fig.~\ref{fig:biodex}B) more accurately captured knee flexion (Fig.~\ref{fig:biodex}C) with a maximum measurement of approximately 200 \unit{\pascal} ($R^2=0.95$). Meanwhile, the pad below the knee (Fig.~\ref{fig:biodex}E) more accurately captured knee extension (Fig.~\ref{fig:biodex}G) with a range of 60 Pa, $R^2=0.75$). In contrast, extension above the knee (Fig.~\ref{fig:biodex}D) and flexion below the knee (Fig.~\ref{fig:biodex}F) yielded weak relationships ($R^2=0.085$ and $0.54$, respectively). This was not surprising because the force applied by the shank during extension pushed into the pad, whereas in flexion, the force pushes away. Conversely, during extension, the thigh applied forces downward into the seat, which the pad would not register as accurately; however, in flexion, the thigh applies forces into the pad with measurable deformation. Additionally, the pad below the knee was adjacent to the dynamometer mount on the same segment, thus registering only a portion of segment loading and likely distorting the measured pressure signal. This placement could also explain the large pressure variation near zero torque, as local contact pressure and limb stabilization with the dynamometer mount may have exacerbated loading-unloading hysteresis of the pad-soft-tissue interface. Future studies should integrate FI pads directly at the dynamometer mount or primary human–machine interaction site to eliminate the confounding effect of pad placement. Nonetheless, these results suggest strategic pad design can accurately capture user-device interactions. 


\section{Results}
To evaluate FI's capture of human-device interaction during movement with varying loads, we present two test cases featuring cyclic motion of upper- and lower-limbs. 

\subsection{Dynamic and Isometric Bicep Curl Experiment}
\label{sec:arms}
\begin{figure*}[t]
  \centering
  \includegraphics[width=0.99\linewidth]{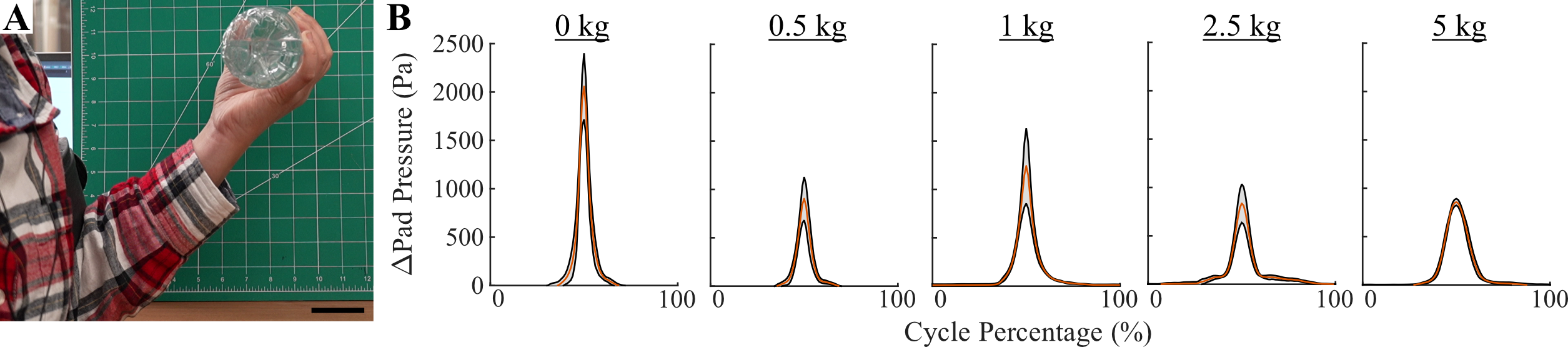}
  \caption{(A) Bicep curls from 90\unit{\degree} to full elbow flexion back to 90\unit{\degree} were repeated with varying load held in the hand ranging 0--4.54 \unit{\kilo\gram} (scale bar 5 \unit{\centi\metre}). (B) Pressure data normalized in time by cycle percentage is shown for each load (orange and black lines are mean ± 1 SD, respectively).}
  \label{fig:armFull}
  \vspace{-1mm}
\end{figure*}

\begin{figure}[t]
  \centering
  \includegraphics[width=0.99\linewidth]{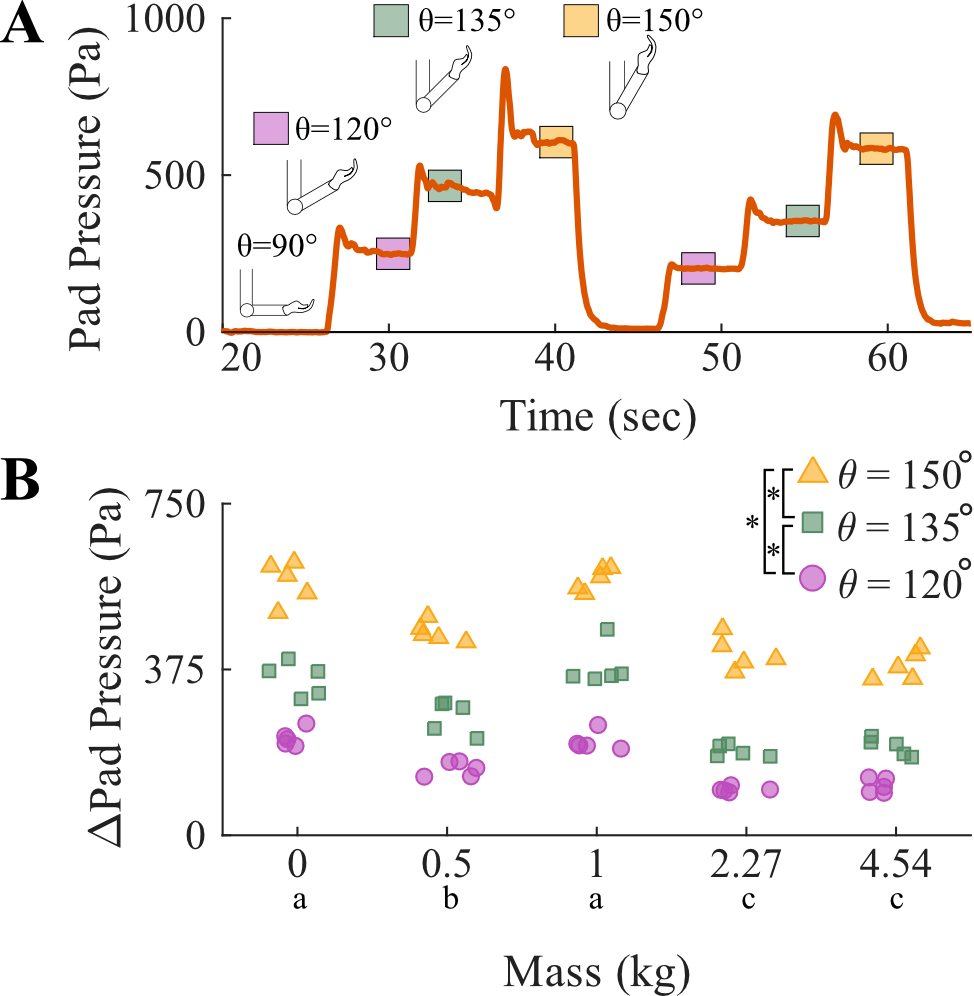}
  \caption{(A) Two exemplary cycles of step-wise isometric bicep curls at three elbow angles while holding 1\unit{\kilo\gram}. (B) Mass held vs steady-state pad pressure at each angle (120\unit{\degree}, 135\unit{\degree}, and 150\unit{\degree} purple circle, green square, and yellow triangle, respectively). Pressure consistently significantly increased with higher elbow angle. (*$p<$0.001). Masses had non-monotonic effects, with different letters indicating significant differences ($\alpha<$0.05).}
  \label{fig:armStep}
  \vspace{-5mm}
\end{figure}

Elbow flexion experiments were conducted with the sensing pad  strapped on the \textit{biceps brachii} of a single healthy individual (female, 30 years). Elbow flexion was first repeated from 90\unit{\degree} to full flexion and then returning back to 90\unit{\degree} five times (cycle time 4 \unit{\second}) while holding masses of 0, 0.5, 1, 2.27, and 4.54 \unit{\kilo\gram}. Thereafter, for each mass, the user conducted another 5 cycles; within each cycle, they paused at $\approx120\unit{\degree}$, 135\unit{\degree}, and 150\unit{\degree} for 4 \unit{\second} each. In the full cycle dataset, pressure data were filtered and offset, each cycle was normalized (linear interpolation, 100 data points), and the mean and SD across cycles were computed for each mass (Fig.~\ref{fig:armFull}). To extract steady-state pressure in the stepwise cycle dataset, at each position, a 2 \unit{\second} region with the lowest coefficient of variation within each cycle was isolated and averaged (Fig.~\ref{fig:armStep}A). A 2-way analysis of variance assessed effects of elbow angle, mass held, and their interaction, with post-hoc multiple comparisons conducted for any significant effects (Tukey's LSD).

In the dynamic full cycles of elbow flexion (Fig.~\ref{fig:armFull}), the pad pressure exhibited a clear correlation between pressure and elbow angle across all masses. Pressure sharply increased then decreased with remarkable repeatability across cycles. Unexpectedly, non-monotonic peak pad pressure was observed with increases in the mass held in the hand. Highest peak pressure of 2000 \unit{\pascal} was seen with no load, followed by 1200 \unit{\pascal} with 1 \unit{\kilo\gram}, and lower peak pressures ranging between 600 to 800 \unit{\pascal} for 0.5, 2.27, and 4.45 \unit{\kilo\gram}. The non-monotonic behavior at 0.5 \unit{\kilo\gram} may reflect variability in limb-pad contact mechanics across loads in an unconstrained task, and warrants further investigation with more participants.

These trends persisted in the stepwise isometric bicep curls (Fig.~\ref{fig:armStep}). Elbow angle strongly influenced pad pressure across all loads ($F(2,60)=772.5$, $p<0.001$), with pressure increasing with flexion. There was a significant effect of load ($F(4,60)=103.8$, $p<0.001$) where a general reduction in pressure was observed for increasing mass at each angle. However, this was not strictly monotonic as pressures in 0.5 \unit{\kilo\gram} were significantly lower than 0 and 1 \unit{\kilo\gram} but significantly greater than 2.27 and 4.54 \unit{\kilo\gram}. A significant angle-load interaction was present ($F(8,60)=2.87$, $p=0.009$), with more flexion tending to amplify pressure differences at higher loads. Together with the dynamometer results, these findings underscore the importance of strategically designing the pad geometry and placement to capture meaningful deformation under relevant mechanical loads. 


\subsection{Squats Experiment}
\label{sec:squats}
\begin{figure}[t]
  \centering
  \includegraphics[width=3.5in, keepaspectratio]{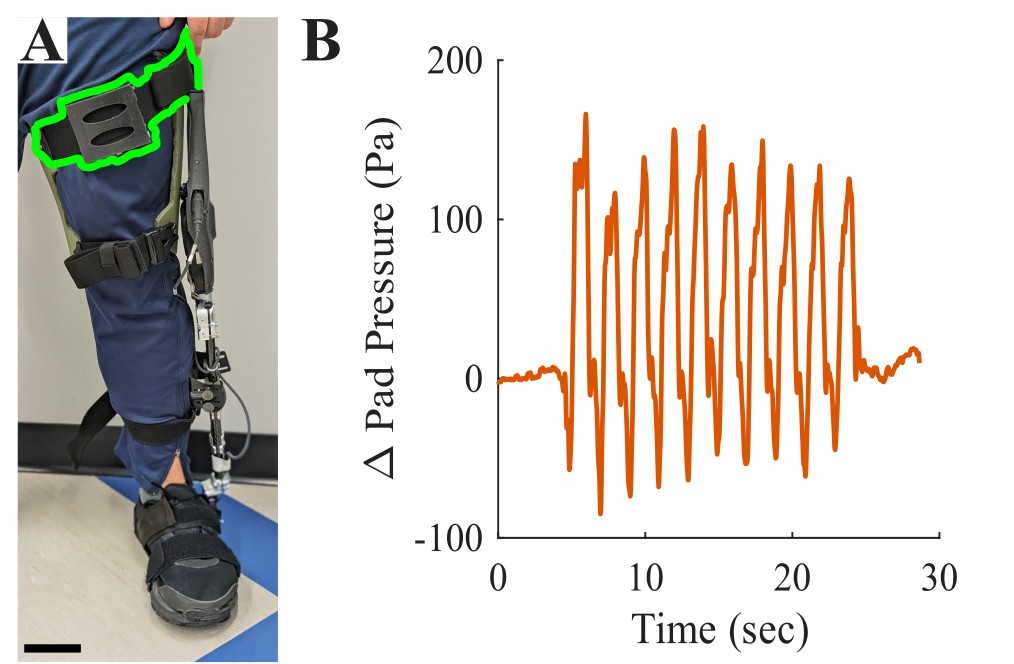}
  \caption{(A) Fluidically innervated pads were integrated into a robotic exoskeleton \cite{devine2025robotic}. Scale bar is 5 \unit{\centi\metre}. (B) Change in pad pressure in 10 squats (2 \unit{\second}/cycle), with max positive pressure near lowest portion of a squat.
}
  \label{fig:squats}
  \vspace{-5mm}
\end{figure}

In the lower-limb experiment, we integrated a 3D printed pad into the upper-thigh strap of a robotic knee exoskeleton (Fig.~\ref{fig:squats}A). The exoskeleton has a single actuator at the knee joint, underfoot FSRs, and a knee-ankle-foot-orthosis (details in \cite{devine2025robotic}). To isolate mechanical interaction from volitional movement between the user and device, the robot was not powered. While wearing the device, a single healthy individual (male, 21 years) stood still for 4 \unit{\second}, and then repeated a squat to $\approx 90\unit{\degree}$ of knee flexion 10 times (2 \unit{\second} cycle time) with a metronome cuing each half cycle. Pad pressure was offset and filtered as done in Sec.~\ref{sec:dynamometer}.

During squatting, pad pressure exhibited a cyclic pattern that reflected task dynamics (Fig.~\ref{fig:squats}B), with maximum pressure at the most flexed part of the squat. This result aligns with the above-knee flexion condition in Sec.~\ref{sec:dynamometer}, suggesting thigh deformation can be captured in both isometric and dynamic conditions. This result is also consistent with upper-limb results showing pressure dependence on joint angle. While these results suggest the system can be integrated with a wearable robot, future extended testing with the robot powered are needed to evaluate user comfort, hysteresis, potential signal drift, and overall ability to capture volitional user interaction and applied torques by the device. 


\section{Discussion}
We presented a wearable pad sensorized using FI that captures interaction forces via pressure changes. The sensor was evaluated under progressively less constrained conditions: (i) benchtop compression testing, (ii) dynamometer-mediated isometric knee exercises, (iii) stepwise isometric and dynamic bicep curls, and (iv) integration into an unpowered lower-extremity exoskeleton during repeated squats. The mechanical testing results validate the capability of FI to reliably capture direct interaction forces. The tight correlation between force and pressure are a significant improvement over traditional FSRs and their non-linear force response \cite{velasquez2019error}. We also demonstrate that strategically placed pads can accurately capture human torque output during isometric knee flexion and extension. Results from unconstrained arm movements demonstrate pad sensitivity to external load and arm angle. Both factors are related to the net moment output by the human user. Similarly, during squatting, pad pressure was strongly linked with knee flexion angle, which also correlates with the required knee moment to provide anti-gravity body weight support.

Collectively, these findings provide impetus for continued development and eventual integration of FI pads into wearable robotic devices. Compared to prior FI work in robotic grippers \cite{zhang2024embedded, shang2025forte}, our results demonstrate this sensing modality translates effectively to the more variable human-robot interface. Still, greater signal variability due to the inherent unconstrained and complex nature of human dynamics were present, leaving several remaining design challenges. First, multiple factors (i.e. precise force direction relative to pad, limb angle, external load, strap tightness) may have significant effects on measured pad pressure; future work should examine the interaction between these factors in more detail. Integration of these pads into a multimodal system that also tracks kinematics using an IMU or other sensors may lead to improved interpretation of pressure readings in terms of human effort by accounting for limb dynamics. It is likely deployment of miniaturized pressure pads, or distinct pressure sensing channels within a larger surface area, would enhance characterization of the human-machine interface. 

Overall, our initial results show promise for FI sensorized pads to accurately quantify interaction force in wearable robotic applications with high resolution and low latency. In the future, these force measurements could provide useful information during device use, such as user movement intent and force output, which could be integrated into the control loop to adapt and optimize robotically applied forces in real time \cite{divekar2024versatile} and over longer time scales \cite{zhang2017human}. Direct force measurement could also enhance user comfort by minimizing unwanted interaction forces in real-time and by quantifying pressure distributions to inform future device designs.   

\section*{Acknowledgment}

This research was supported by the Intramural Research Program at the National Institutes of Health (NIH), the Texas Robotics Industrial Affiliates Program, and the Schmidt Science Fellows program. The contributions of the NIH authors are considered Works of the United States Government. The findings and conclusions presented in this paper are those of the authors and do not necessarily reflect the views of the NIH or the U.S. Department of Health and Human Services. 

\balance
\bibliographystyle{ieeetr}
\bibliography{bibliography}

@book{winter2009biomechanics,
  title={Biomechanics and motor control of human movement},
  author={Winter, David A},
  year={2009},
  publisher={John wiley \& sons}
}

@article{tamez2015real,
  title={Real-time strap pressure sensor system for powered exoskeletons},
  author={Tamez-Duque, Jes{\'u}s and Cobian-Ugalde, Rebeca and Kilicarslan, Atilla and Venkatakrishnan, Anusha and Soto, Rogelio and Contreras-Vidal, Jose Luis},
  journal={Sensors},
  volume={15},
  number={2},
  pages={4550--4563},
  year={2015},
  publisher={Multidisciplinary Digital Publishing Institute}
}

@inproceedings{bessler2019prototype,
  title={Prototype measuring device for assessing interaction forces between human limbs and rehabilitation robots-a proof of concept study},
  author={Bessler, Jule and Schaake, Leendert and Kelder, Roy and Buurke, Jaap H and Prange-Lasonder, Gerdienke B},
  booktitle={2019 IEEE 16th International Conference on Rehabilitation Robotics (ICORR)},
  pages={1109--1114},
  year={2019},
  organization={IEEE}
}

@article{lenzi2011measuring,
  title={Measuring human--robot interaction on wearable robots: A distributed approach},
  author={Lenzi, Tommaso and Vitiello, Nicola and De Rossi, Stefano Marco Maria and Persichetti, Alessandro and Giovacchini, Francesco and Roccella, Stefano and Vecchi, Fabrizio and Carrozza, Maria Chiara},
  journal={Mechatronics},
  volume={21},
  number={6},
  pages={1123--1131},
  year={2011},
  publisher={Elsevier}
}

@article{langlois2020investigating,
  title={Investigating the effects of strapping pressure on human-robot interface dynamics using a soft robotic cuff},
  author={Langlois, Kevin and Rodriguez-Cianca, David and Serrien, Ben and De Winter, Joris and Verstraten, Tom and Rodriguez-Guerrero, Carlos and Vanderborght, Bram and Lefeber, Dirk},
  journal={IEEE transactions on Medical Robotics and Bionics},
  volume={3},
  number={1},
  pages={146--155},
  year={2020},
  publisher={IEEE}
}

@article{wang2021differential,
  title={Differential soft sensor-based measurement of interactive force and assistive torque for a robotic hip exoskeleton},
  author={Wang, Sun’an and Zhang, Binquan and Yu, Zhenyuan and Yan, Yu’ang},
  journal={Sensors},
  volume={21},
  number={19},
  pages={6545},
  year={2021},
  publisher={MDPI}
}

@article{langlois2021integration,
  title={Integration of 3D printed flexible pressure sensors into physical interfaces for wearable robots},
  author={Langlois, Kevin and Roels, Ellen and Van De Velde, Gabri{\"e}l and Espadinha, Cl{\'a}udia and Van Vlerken, Christopher and Verstraten, Tom and Vanderborght, Bram and Lefeber, Dirk},
  journal={Sensors},
  volume={21},
  number={6},
  pages={2157},
  year={2021},
  publisher={MDPI}
}

@article{divekar2024versatile,
  title={A versatile knee exoskeleton mitigates quadriceps fatigue in lifting, lowering, and carrying tasks},
  author={Divekar, Nikhil V and Thomas, Gray C and Yerva, Avani R and Frame, Hannah B and Gregg, Robert D},
  journal={Science Robotics},
  volume={9},
  number={94},
  pages={eadr8282},
  year={2024},
  publisher={American Association for the Advancement of Science}
}

@article{zhang2017human,
  title={Human-in-the-loop optimization of exoskeleton assistance during walking},
  author={Zhang, Juanjuan and Fiers, Pieter and Witte, Kirby A and Jackson, Rachel W and Poggensee, Katherine L and Atkeson, Christopher G and Collins, Steven H},
  journal={Science},
  volume={356},
  number={6344},
  pages={1280--1284},
  year={2017},
  publisher={American Association for the Advancement of Science}
}

@inproceedings{rathore2016quantifying,
  title={Quantifying the human-robot interaction forces between a lower limb exoskeleton and healthy users},
  author={Rathore, Ashish and Wilcox, Matthew and Ramirez, Dafne Zuleima Morgado and Loureiro, Rui and Carlson, Tom},
  booktitle={2016 38th Annual International Conference of the IEEE Engineering in Medicine and Biology Society (EMBC)},
  pages={586--589},
  year={2016},
  organization={IEEE}
}

@article{choi2018compact,
  title={Compact hip-force sensor for a gait-assistance exoskeleton system},
  author={Choi, Hyundo and Seo, Keehong and Hyung, Seungyong and Shim, Youngbo and Lim, Soo-Chul},
  journal={Sensors},
  volume={18},
  number={2},
  pages={566},
  year={2018},
  publisher={MDPI}
}

@book{rosenWearable2019,
  title = {Wearable {{Robotics}}: {{Systems}} and {{Applications}}},
  shorttitle = {Wearable {{Robotics}}},
  author = {Rosen, Jacob},
  year = {2019},
  month = nov,
  publisher = {Academic Press},
  isbn = {978-0-12-814660-6},
  langid = {english}
}

@inproceedings{lunenburgerBiofeedback2004,
  title = {Biofeedback in Gait Training with the Robotic Orthosis {{Lokomat}}},
  booktitle = {The 26th {{Annual International Conference}} of the {{IEEE Engineering}} in {{Medicine}} and {{Biology Society}}},
  author = {Lunenburger, L. and Colombo, G. and Riener, R. and Dietz, V.},
  year = {2004},
  month = sep,
  volume = {2},
  pages = {4888--4891},
  doi = {10.1109/IEMBS.2004.1404352},
  pmid = {17271408}
}

@article{luoExperimentfree2024,
  title = {Experiment-Free Exoskeleton Assistance via Learning in Simulation},
  author = {Luo, Shuzhen and Jiang, Menghan and Zhang, Sainan and Zhu, Junxi and Yu, Shuangyue and Dominguez Silva, Israel and Wang, Tian and Rouse, Elliott and Zhou, Bolei and Yuk, Hyunwoo and Zhou, Xianlian and Su, Hao},
  year = {2024},
  month = jun,
  journal = {Nature},
  volume = {630},
  number = {8016},
  pages = {353--359},
  issn = {1476-4687},
  doi = {10.1038/s41586-024-07382-4},
  urldate = {2025-02-06},
  copyright = {2024 The Author(s), under exclusive licence to Springer Nature Limited},
  langid = {english},
  pmcid = {PMC11344585}
}

@article{molinaroTaskagnostic2024,
  title = {Task-Agnostic Exoskeleton Control via Biological Joint Moment Estimation},
  author = {Molinaro, Dean D. and Scherpereel, Keaton L. and Schonhaut, Ethan B. and Evangelopoulos, Georgios and Shepherd, Max K. and Young, Aaron J.},
  year = {2024},
  month = nov,
  journal = {Nature},
  volume = {635},
  number = {8038},
  pages = {337--344},
  issn = {1476-4687},
  doi = {10.1038/s41586-024-08157-7},
  urldate = {2025-02-06},
  copyright = {2024 The Author(s), under exclusive licence to Springer Nature Limited},
  langid = {english},
  pmid = {39537888}
}

@inproceedings{leeControlling2016,
  title = {Controlling Negative and Positive Power at the Ankle with a Soft Exosuit},
  booktitle = {2016 {{IEEE International Conference}} on {{Robotics}} and {{Automation}} ({{ICRA}})},
  author = {Lee, Sangjun and Crea, Simona and Malcolm, Philippe and Galiana, Ignacio and Asbeck, Alan and Walsh, Conor},
  year = {2016},
  month = may,
  pages = {3509--3515},
  doi = {10.1109/ICRA.2016.7487531}
}

@article{babicChallenges2021,
  title = {Challenges and Solutions for Application and Wider Adoption of Wearable Robots},
  author = {Babi{\v c}, Jan and Laffranchi, Matteo and Tessari, Federico and Verstraten, Tom and Novak, Domen and {\v S}arabon, Nejc and Ugurlu, Barkan and Peternel, Luka and Torricelli, Diego and Veneman, Jan F.},
  year = {2021},
  month = jan,
  journal = {Wearable Technologies},
  volume = {2},
  pages = {e14},
  issn = {2631-7176},
  doi = {10.1017/wtc.2021.13},
  urldate = {2026-01-14},
  langid = {english}
}

@article{zhuSoft2022,
  title = {Soft, {{Wearable Robotics}} and {{Haptics}}: {{Technologies}}, {{Trends}}, and {{Emerging Applications}}},
  shorttitle = {Soft, {{Wearable Robotics}} and {{Haptics}}},
  author = {Zhu, Mengjia and Biswas, Shantonu and Dinulescu, Stejara Iulia and Kastor, Nikolas and Hawkes, Elliot Wright and Visell, Yon},
  year = {2022},
  month = feb,
  journal = {Proceedings of the IEEE},
  volume = {110},
  number = {2},
  pages = {246--272},
  issn = {1558-2256},
  doi = {10.1109/JPROC.2021.3140049},
  urldate = {2026-01-14}
}

@article{velasquez2019error,
  title={Error compensation in force sensing resistors},
  author={Vel{\'a}squez, Elkin I Guti{\'e}rrez and G{\'o}mez, V{\'\i}ctor and Paredes-Madrid, Leonel and Colorado, Henry A},
  journal={Sensing and Bio-Sensing Research},
  volume={26},
  pages={100300},
  year={2019},
  publisher={Elsevier}
}

@article{adamczyk2023wearable,
  title={Wearable sensing for understanding and influencing human movement in ecological contexts},
  author={Adamczyk, Peter Gabriel and Harper, Sara E and Reiter, Alex J and Roembke, Rebecca A and Wang, Yisen and Nichols, Kieran M and Thelen, Darryl G},
  journal={Current opinion in biomedical engineering},
  volume={28},
  pages={100492},
  year={2023},
  publisher={Elsevier}
}

@article{truby2022fluidic,
  title={Fluidic innervation sensorizes structures from a single build material},
  author={Truby, Ryan L and Chin, Lillian and Zhang, Annan and Rus, Daniela},
  journal={Science advances},
  volume={8},
  number={31},
  pages={eabq4385},
  year={2022},
  publisher={American Association for the Advancement of Science}
}

@inproceedings{zhang2024embedded,
  title={Embedded air channels transform soft lattices into sensorized grippers},
  author={Zhang, Annan and Chin, Lillian and Tong, Daniel L and Rus, Daniela},
  booktitle={2024 IEEE International Conference on Robotics and Automation (ICRA)},
  pages={5264--5270},
  year={2024},
  organization={IEEE}
}

@ARTICLE{shang2025forte,
  author={Shang, Siqi and Seo, Mingyo and Zhu, Yuke and Chin, Lillian},
  journal={IEEE Robotics and Automation Letters}, 
  title={FORTE: Tactile Force and Slip Sensing on Compliant Fingers for Delicate Manipulation}, 
  year={2026},
  volume={11},
  number={4},
  pages={4473-4480},
  doi={10.1109/LRA.2026.3662618}}

@article{devine2025robotic,
  title={Robotic Knee Exoskeletons as Assistive and Gait Training Tools in Spina Bifida: A Pilot Study Showing Clinical Feasibility of Two Control Strategies},
  author={Devine, Taylor M and Asante-Otoo, Afua and Alter, Katharine and Damiano, Diane L and Bulea, Thomas C},
  journal={IEEE Transactions on Neural Systems and Rehabilitation Engineering},
  year={2025},
  publisher={IEEE}
}
\end{document}